\documentclass{article}

\usepackage{PRIMEarxiv}

\usepackage[utf8]{inputenc} 
\usepackage[T1]{fontenc}    
\usepackage{hyperref}       
\usepackage{url}            
\usepackage{booktabs}       
\usepackage{amsfonts}       
\usepackage{nicefrac}       
\usepackage{microtype}      
\usepackage{lipsum}
\usepackage{fancyhdr}       
\usepackage{graphicx}       
\usepackage{pifont}
\usepackage{xcolor}         
\usepackage{subcaption}
\usepackage{multirow}
\usepackage[normalem]{ulem}

\usepackage{titlesec}

\graphicspath{{media/}}     

\title{Isharah-Selfie: Continuous Sign Language Recognition Dataset for One-handed Signing}

\author{%
  \parbox{\textwidth}{\centering
    Ahmed Abul Hasanaath$^{1}$, Murtadha Aljubran$^{2}$, Sarah Alyami$^{3}$,
    Muhammad Haris Khan$^{2}$, Hamzah Luqman$^{1}$ \\[4pt]
    {\normalfont\small
    $^{1}$King Fahd University of Petroleum \& Minerals, $^{2}$Mohamed bin Zayed University of Artificial Intelligence,
    $^{3}$Imam Abdulrahman Bin Faisal University} \\[4pt]
    {\normalfont\small \texttt{hluqman@kfupm.edu.sa}}
  }
}

\begin{document}
\maketitle

\begin{center}
\includegraphics[width=0.80\linewidth]{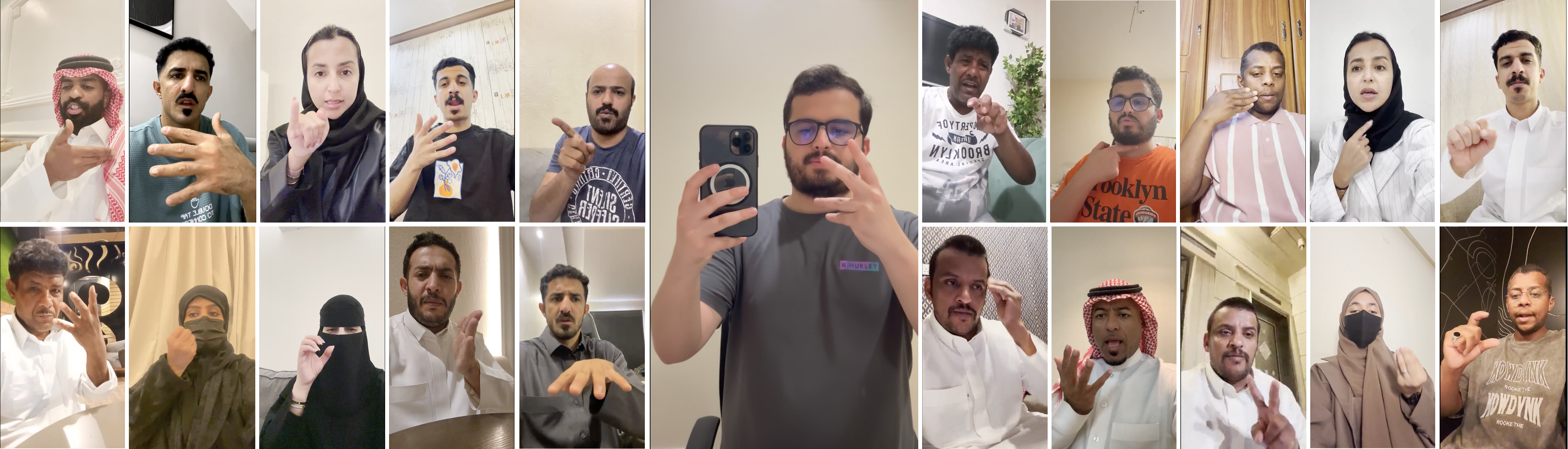}
\captionof{figure}{Samples from the Isharah-Selfie dataset. The central sample depicts a signer recording in selfie mode; the remaining samples illustrate the variety of signers, environments, and recording conditions.}
\label{fig:signers-sample}
\end{center}

\vspace{2em}

\begin{abstract}
Current sign language recognition and translation benchmarks are commonly collected in controlled settings using fixed or studio-style cameras, particularly limiting their ability to reflect how signers communicate in everyday mobile scenarios. As such, this gap becomes increasingly evident when considering the Arabic Sign Language (ArSL), which is a primary sign language used across Arab countries. 
In this work, we introduce Isharah-Selfie, a large-scale ArSL dataset collected entirely using front-facing smartphone cameras under unconstrained selfie conditions. The dataset comprises 8,997 video clips performed by 16 signers across 1,493 unique sentences, covering several domains such as healthcare, education, and transportation. Unlike conventional datasets dominated by fixed-camera, two-handed signing, Isharah-Selfie captures realistic mobile communication characterized by one-handed signing, close-range viewpoints, partial body visibility, camera motion, unconstrained environments, and heterogeneous device resolutions. Each video is annotated with both gloss sequences for continuous sign language recognition (CSLR) and Arabic sentence translations for sign language translation (SLT). Moreover, we define signer-independent and unseen-sentences evaluation protocols to assess generalization across unseen signers and novel sentence compositions. Extensive benchmarks using RGB-based and pose-based data show that current CSLR and SLT methods remain challenged by selfie-style signing conditions, particularly under unseen-sentence evaluation, where compositional generalization remains weak. By releasing Isharah-Selfie, we aim to support research toward robust, user-centered sign language technologies that operate reliably in realistic smartphone-based communication settings. The Isharah-Selfie dataset is available on \url{https://isharah-selfie.github.io/}.
\end{abstract}


\section{Introduction} 
\label{sec:introduction}

Sign language is a natural visual-gestural language that uses hand movements, facial expressions, body posture, and spatial cues to convey meaning. Unlike spoken languages, sign language operates through visual-manual communication and has its own morphology, syntax, and pragmatics~\cite{sandler2006sign, brentari2018sign}. Sign language is shaped by local culture, history, and geography, resulting in many different and distinct sign languages worldwide, such as American, Chinese, and Arabic sign languages. 
Arabic Sign Language (ArSL) is the primary sign language used across Arab countries. Although these countries share a common spoken language, multiple sign language variants exist within local Arab communities, exhibiting minor differences, such as Saudi Sign Language (SSL), Qatari Sign Language (QSL), and Moroccan Sign Language.  
SSL is a variant of ArSL and serves as the primary communication medium within the Saudi Deaf community, which includes over 84,025 Deaf individuals~\cite{DisabilityReport}. Moreover, SSL is also used in other Arab Gulf countries and is the predominant sign language featured in most media across the region.

In recent years, Sign Language Recognition (SLR) and Sign Language Translation (SLT) have emerged as prominent research areas aimed at enhancing accessibility and facilitating inclusive communication between deaf and hearing individuals.
SLR tasks are typically divided into Isolated SLR (ISLR) and Continuous SLR (CSLR), where the latter focuses on recognizing sentences composed of fluid, interconnected signs~\cite{Laines_2023_CVPR, ALYAMI2024103774}. 
CSLR remains challenging due to the absence of clear sign boundaries, co-articulation between consecutive gestures, and variations in signing style~\cite{koulierakis2021sign, athira2022signer}. 
In contrast to CSLR, which outputs recognized sentences as a sequence of glosses, where a gloss is a written representation of a sign that preserves the structure and grammar of the source sign language, SLT extends this task by generating natural spoken or written sentences from sign sequences. This process bridges linguistic and semantic gaps between sign and spoken languages~\cite{Zhou_2023_ICCV}.

While the development of CSLR and SLT systems relies heavily on large annotated datasets, existing resources are typically collected in controlled environments using static camera setups~\cite{Dreuw2009, Camgoz2018, Koller2015, Kagirov2020, Adaloglou2021}. 
Such datasets do not fully capture how deaf individuals naturally communicate in daily life. 
With the increasing use of smartphones for video communication, signers commonly rely on the smartphone's front-facing camera, typically in ``selfie mode'', to communicate with both deaf and hearing people. Moreover, Deaf individuals often employ this recording setup to share their daily activities on social media and video-sharing platforms.
{\color{black}
Our motivation was further informed by observations and discussions with members of the local Deaf community, who confirmed that one-handed signing is used in short, informal smartphone interactions when the other hand holds the device.
}
{\color{black}
This observation is consistent with prior  findings~\cite{cavender2006mobileasl}, in which ASL Deaf participants considered holding a phone while signing feasible for short conversations, while preferring to place the device on a surface for longer interactions.
}
This recording setup differs substantially from the conventional data collection protocols used in most CSLR and SLT datasets. One-handed signing introduces close-range viewpoints, partial body visibility, variable camera distance and angle, dynamic camera motion, and frequent one-handed articulation when the other hand is occupied, such as when holding a smartphone.

Recent work has begun to recognize the importance of smartphone-based signing. For example, the PopSign dataset~\cite{starner2023popsign} introduced a large-scale isolated American Sign Language (ASL) dataset collected using smartphone selfie cameras, focusing specifically on one-handed signing. The findings highlight that one-handed signing is a realistic condition in smartphone-based interactions and that expert signers can adapt signs when communicating through mobile devices. However, PopSign focuses on isolated ASL signs for educational, game-based recognition, whereas CSLR and SLT require modeling full signed sentences with co-articulation, temporal dependencies, and sentence-level linguistic structure. Moreover, smartphone-based continuous signing remains underexplored, particularly for underrepresented languages such as SSL.

To study this natural and underexplored signing behavior, we introduce \textit{Isharah-Selfie}, {\color{black}a new dataset of continuous SSL sentences} recorded by deaf individuals using their smartphone front cameras. 
{\color{black}
The name \textit{Isharah}, means ``sign'' or ``gesture'' in Arabic.
}
This dataset is the first large-scale CSLR dataset collected entirely in handheld, unconstrained selfie-mode conditions. It comprises 1,493 unique sentences, each performed by 16 signers, resulting in 8,997 real-world video clips. As illustrated in Figure~\ref{fig:signers-sample}, the dataset spans a wide range of recording scenarios, reflecting variability in camera angle, framing, motion trajectories, background environments, and illumination. These characteristics naturally produce partial occlusions and dynamic visual changes that are rarely present in existing sign language datasets, making \textit{Isharah-Selfie} a challenging yet highly realistic benchmark.
Moreover, the dataset is linguistically rich as it provides gloss-level annotations for all videos, making it useful for developing CSLR and SLT systems and for studying one-handed adaptations in SSL.

Unlike the original Isharah dataset~\cite{alyami2025isharah}, which primarily features multi-scene, two-handed recording setups, \textit{Isharah-Selfie} captures authentic, self-recorded signing in everyday environments. This shift introduces several challenges for SLR and SLT models, including partial body visibility, significant camera motion, unconstrained lighting, background clutter, and the systematic use of one-handed signing patterns that compress inherently bimanual gestures into a single active hand. 

\begin{figure*}[t]
  \centering
  \includegraphics[width=.7\textwidth]{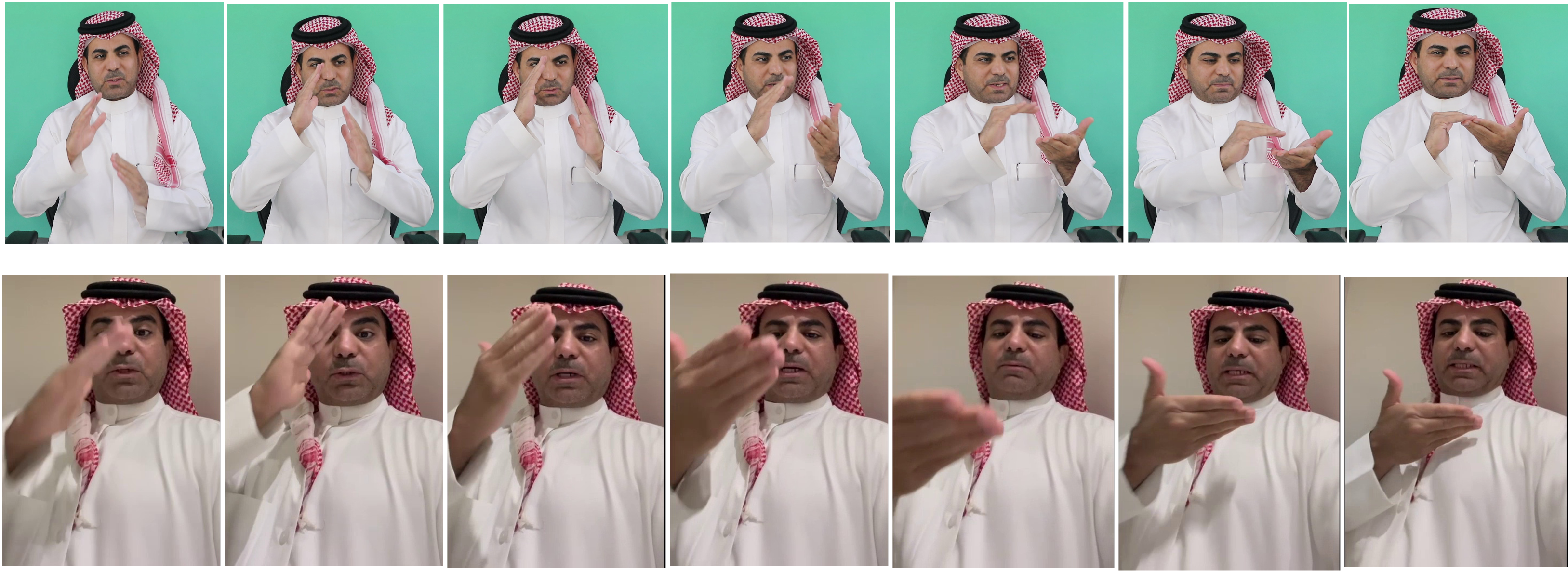}
\caption{Two-handed vs. selfie one-handed signing comparison. The sentence, ``thank you for help,'' is shown under conventional two-handed signing (top) and selfie-based one-handed signing (bottom). 
The selfie setting reduces the signing space, removes the non-dominant hand, and introduces a closer viewpoint, hand--face occlusion, and perspective distortion.
}
  \label{fig:one_hand_signing_challenges}
\end{figure*}

As illustrated in Figure~\ref{fig:one_hand_signing_challenges}, using the same sentence signed in both settings, the selfie version exhibits reduced signing space, absence of the non-dominant hand, increased hand--face occlusion, and stronger perspective distortion due to close-range mobile capture. Addressing these complexities is essential for developing user-centric, context-aware CSLR and SLT systems capable of generalizing beyond controlled laboratory conditions.

\section{Related Work} 
\label{sec:related_work}

Early CSLR datasets were often small and collected under constrained conditions. RWTH-BOSTON-104 \cite{Dreuw2009} contains low-resolution videos from only three signers, while later resources such as SIGNUM \cite{VonAgris2007} and RWTH-PHOENIX-Weather \cite{Forster2012} expanded the scale of German Sign Language (DGS) data but remained largely controlled or domain-specific. RWTH-PHOENIX-2014 \cite{Koller2015,forster-etal-2014-extensions} further standardized CSLR evaluation and remains a widely used benchmark. Other CSLR datasets have broadened language coverage, including CSL \cite{Huang2018}, CSL-Daily \cite{Zhou2021}, GSL \cite{Adaloglou2021}, ArabSign \cite{luqman2022arabsign}, JUMLA-QSL-22 \cite{GhoulJumla2023}, FluentSigners-50 \cite{Mukushev2022}, and TVB-HKSL-News \cite{niu-etal-2024-hong}. However, many of these datasets are limited by controlled recording setups, domain-specific content, limited signer diversity, or small sentence vocabularies.

SLT datasets pair sign videos with spoken or written language annotations. Influential resources include RWTH-PHOENIX-2014T \cite{Camgoz2018}, BOBSL \cite{albanie2021bbc}, How2Sign \cite{Duarte2021}, OpenASL \cite{shi2022open}, YouTube-ASL \cite{uthus2023youtube}, Auslan-Daily \cite{shen2023auslan}, and YouTube-SL-25 \cite{tanzer2024youtube}. These datasets have advanced video-to-text translation, especially for European and American sign languages, and many are collected from broadcast or web videos. Large linguistic corpora such as the BSL Corpus \cite{schembri2013building} and Public DGS Corpus \cite{hanke2020extending} also provide valuable sign language resources, although their annotation structure and segmentation protocols may differ from standard CSLR and SLT benchmarks.

Despite this progress, few datasets capture self-recorded signing where signers interact directly with mobile devices in everyday environments. As summarized in Table~\ref{tab:sl_datasets}, existing resources are commonly collected in studio, broadcast, web-video, or professionally framed settings, while smartphone selfie signing introduces additional challenges such as camera motion, close-range framing, partial body visibility, heterogeneous resolutions, and one-handed articulation. The original Isharah dataset \cite{alyami2025isharah} introduced a large-scale ArSL resource with 30,000 smartphone-recorded videos and both gloss and Arabic text annotations, but it primarily focuses on multi-scene recording rather than selfie-based one-handed signing. To address this gap, we introduce \textit{Isharah-Selfie}, a new SSL dataset recorded using front-facing smartphone cameras to support more realistic CSLR and SLT benchmarking under mobile self-recorded conditions. Additional discussion on CSLR and SLT datasets is provided in Appendix \ref{appen_RL}.

\begin{table*}[t]
\centering
\caption{Summary of publicly available sign language datasets. Datasets without gloss annotations are listed first and are primarily used for gloss-free SLT, whereas gloss-based datasets provide gloss annotations for each sign language sentence and are used for CSLR and gloss-based SLT. \textbf{SL}: sign language; \textbf{Domain}: single (S) or multiple domains (M); \textbf{Phone Cam.}: signs captured using smartphone;  \textbf{Unconstr.}: unconstrained recording environment; \textbf{1-Handed}: one-handed signing.}
\label{tab:sl_datasets}
\footnotesize
\setlength{\tabcolsep}{4pt}
\resizebox{\linewidth}{!}{
\begin{tabular}{@{}lccccccccccc@{}}
\toprule
Dataset & Year & SL &
Gloss Vocab. & Text Vocab. &
Signers & Duration (h) & Domain & Phone Cam.  &
Unconstr. & 1-Handed & Motion \\
\midrule

\multicolumn{12}{c}{\textbf{Gloss-Free Sign Language Datasets}} \\
\midrule

KETI~\cite{ko2019neuralsignlanguagetranslation}
& 2019 & KSL &  & 500 & 14 & 27.9 & S & \textcolor{red}{\ding{55}}
& \textcolor{red}{\ding{55}} & \textcolor{red}{\ding{55}} & \textcolor{red}{\ding{55}} \\

BOBSL~\cite{albanie2021bbc}
& 2021 & BSL &  & 78K & 39 & 1467 & M & \textcolor{red}{\ding{55}}
& \textcolor{blue}{\ding{51}} & \textcolor{red}{\ding{55}} & \textcolor{red}{\ding{55}} \\

SWISSTXT-Weather~\cite{camgoz2021content4all}
& 2021 & DSGS &  & 1248 & 12 & 1 & S & \textcolor{red}{\ding{55}}
& \textcolor{blue}{\ding{51}} & \textcolor{red}{\ding{55}} & \textcolor{red}{\ding{55}} \\

VRT-NEWS~\cite{camgoz2021content4all}
& 2021 & VGT &  & 6875 &  & 9 & S & \textcolor{red}{\ding{55}}
& \textcolor{blue}{\ding{51}} & \textcolor{red}{\ding{55}} & \textcolor{red}{\ding{55}} \\

How2Sign~\cite{Duarte2021}
& 2021 & ASL &  & 16K & 11 & 80 & M & \textcolor{red}{\ding{55}}
& \textcolor{red}{\ding{55}} & \textcolor{red}{\ding{55}} & \textcolor{red}{\ding{55}} \\

SpreadTheSign~\cite{yin2022mlslt}
& 2022 & Multi &  & 16700 & 79 & 14 & M & \textcolor{red}{\ding{55}}
& \textcolor{red}{\ding{55}} & \textcolor{red}{\ding{55}} & \textcolor{red}{\ding{55}} \\

OpenASL~\cite{shi2022open}
& 2022 & ASL &  & 3300 & 200 & 288 & M & \textcolor{red}{\ding{55}}
& \textcolor{blue}{\ding{51}} & \textcolor{red}{\ding{55}} & \textcolor{red}{\ding{55}} \\

Auslan-Daily~\cite{shen2023auslan}
& 2023 & Auslan &  & 14K &  & 45 & M & \textcolor{red}{\ding{55}}
& \textcolor{blue}{\ding{51}} & \textcolor{red}{\ding{55}} & \textcolor{red}{\ding{55}} \\

YouTube-ASL~\cite{uthus2023youtube}
& 2023 & ASL &  & 60K & 2519+ & 984 & M & \textcolor{red}{\ding{55}}
& \textcolor{blue}{\ding{51}} & \textcolor{red}{\ding{55}} & \textcolor{red}{\ding{55}} \\

YouTube-SL-25~\cite{tanzer2024youtube}
& 2024 & Multi &  &  & 3000+ & 2980 & M & \textcolor{red}{\ding{55}}
& \textcolor{blue}{\ding{51}} & \textcolor{red}{\ding{55}} & \textcolor{red}{\ding{55}} \\

CSL-News~\cite{li2025unisign}
& 2025 & CSL &  & 5K &  & 1985 & S & \textcolor{red}{\ding{55}} 
& \textcolor{red}{\ding{55}} & \textcolor{red}{\ding{55}} & \textcolor{red}{\ding{55}} \\

\midrule
\multicolumn{12}{c}{\textbf{Gloss-based Sign Language Datasets}} \\
\midrule

PHOENIX-Weather~\cite{Forster2012}
& 2012 & DGS &  &  & 7 & 3.75 & S & \textcolor{red}{\ding{55}}
& \textcolor{red}{\ding{55}} & \textcolor{red}{\ding{55}} & \textcolor{red}{\ding{55}} \\

BSL Corpus~\cite{schembri2013building}
& 2013 & BSL &  &  & 249 &  & M & \textcolor{red}{\ding{55}} 
& \textcolor{red}{\ding{55}} & \textcolor{red}{\ding{55}} & \textcolor{red}{\ding{55}} \\

PHOENIX-2014~\cite{forster-etal-2014-extensions}
& 2014 & DGS & 2,048 &  & 9 & 11 & S & \textcolor{red}{\ding{55}} 
& \textcolor{red}{\ding{55}} & \textcolor{red}{\ding{55}} & \textcolor{red}{\ding{55}} \\

Phoenix-2014T~\cite{Camgoz2018}
& 2018 & DGS & 1066 & 2887 & 9 & 11 & S & \textcolor{red}{\ding{55}} 
& \textcolor{red}{\ding{55}} & \textcolor{red}{\ding{55}} & \textcolor{red}{\ding{55}} \\

Public DGS Corpus~\cite{hanke2020extending}
& 2020 & DGS &  &  & 327 & 50 & S & \textcolor{red}{\ding{55}}
& \textcolor{red}{\ding{55}} & \textcolor{red}{\ding{55}} & \textcolor{red}{\ding{55}} \\

GSL Dataset~\cite{adaloglou2020comprehensive}
& 2020 & GSL & 310 &  & 7 & 9.59 & S & \textcolor{red}{\ding{55}}
& \textcolor{red}{\ding{55}} & \textcolor{red}{\ding{55}} & \textcolor{red}{\ding{55}} \\

BSL-1K~\cite{albanie2020bsl1k}
& 2020 & BSL &  &  & 1000+ & 273 & M & \textcolor{red}{\ding{55}} 
& \textcolor{blue}{\ding{51}} & \textcolor{red}{\ding{55}} & \textcolor{red}{\ding{55}} \\

CSL-Daily~\cite{Zhou2021}
& 2021 & CSL & 2000 & 2300 & 10 & 23 & S & \textcolor{red}{\ding{55}} 
& \textcolor{red}{\ding{55}} & \textcolor{red}{\ding{55}} & \textcolor{red}{\ding{55}} \\

ArabSign~\cite{luqman2022arabsign}
& 2022 & ArSL & 95 &  & 6 & 10 & S & \textcolor{red}{\ding{55}} 
& \textcolor{red}{\ding{55}} & \textcolor{red}{\ding{55}} & \textcolor{red}{\ding{55}} \\

JUMLA-QSL-22~\cite{GhoulJumla2023}
& 2023 & ArSL (QSL) & 201 &  & 7 & 28 & S & \textcolor{red}{\ding{55}} 
& \textcolor{red}{\ding{55}} & \textcolor{red}{\ding{55}} & \textcolor{red}{\ding{55}} \\ 

Isharah~\cite{alyami2025isharah}
& 2025 & ArSL (SSL) & 1132 & 2700 & 18 & 43.19 & M & \textcolor{blue}{\ding{51}}
& \textcolor{blue}{\ding{51}} & \textcolor{red}{\ding{55}} & \textcolor{red}{\ding{55}} \\

\specialrule{0.08em}{2pt}{2pt}

\textbf{Isharah-Selfie (Ours)}
& \textbf{2026} & \textbf{ArSL (SSL)}
& \textbf{961} & \textbf{2133}
& \textbf{16} & \textbf{12.5} & \textbf{M} & \textcolor{blue}{\ding{51}}
& \textcolor{blue}{\ding{51}} & \textcolor{blue}{\ding{51}} & \textcolor{blue}{\ding{51}} \\

\bottomrule
\end{tabular}
}
\end{table*}

\section{Dataset Development}
 \label{sec_dataset}
 
\begin{figure}[h]
\centering
\includegraphics[width=\linewidth]{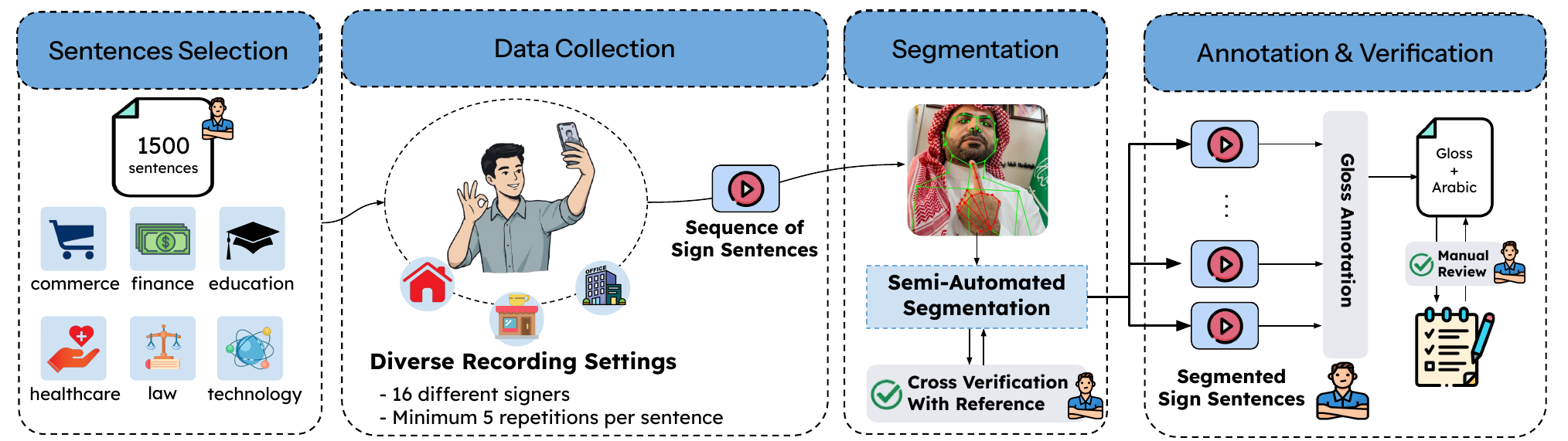}
\caption{Overview of the main stages of the \textit{Isharah-Selfie} dataset development pipeline.}
\label{fig:dataset_pipeline}
\end{figure}

This section describes the development stages of the \textit{Isharah-Selfie} dataset, covering the main stages of sentence design, data collection, video segmentation, annotation, verification, and benchmark partitioning. It further provides a summary of the dataset statistics and evaluation protocols. Figure~\ref{fig:dataset_pipeline} shows an overview of the dataset construction pipeline.

\begin{figure*}[h]
  \centering
  \includegraphics[width=.8\textwidth]{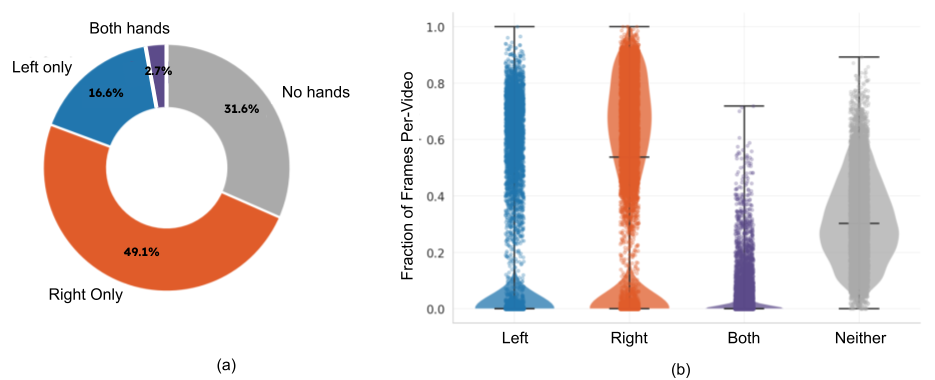}
  \caption{Frame-level hand state distribution (a) and per-video hand presence ratios (b).}
  \label{fig:frame_state_distribution}
\end{figure*}

\subsection{Sentences and Signers}
\vspace{1mm}\noindent\textbf{Sentences Selection. }  
The \textit{Isharah-Selfie} dataset contains 1,493 unique sentences designed in collaboration with domain experts to encompass seven key areas of everyday communication: education (12.2\%), healthcare (24.2\%), justice/courts (12\%), public transportation and airports (24.1\%), emergency services (7.8\%), finance (12.1\%), and general daily interactions (7.6\%). All sentences underwent iterative refinement and cross-validation by certified sign language experts to ensure that each utterance is semantically faithful and naturally signable.

\vspace{1mm}\noindent\textbf{Signers Selection. }
The proposed dataset includes recordings from 16 distinct signers. All signers provided informed consent for the use of their recorded video clips for research purposes. The signers were compensated in accordance with local standards for their participation in the data collection process.
Each signer involved in this dataset recorded approximately 500 sentences in ArSL, specifically using the SSL dialect. The signers comprise a mix of male and female participants, with most aged between 31 and 40 years. About half of the signers are Deaf, while the rest include hard-of-hearing and hearing ArSL translators, ensuring balanced accessibility representation. 
Moreover, some female signers cover their faces using masks or a Niqab, a form of traditional attire commonly worn in many Arab countries. This variation in signer appearance and face visibility enhances the diversity of the dataset and makes it more suitable for developing robust, real-time SLR systems.

\subsection{Dataset Collection}
The dataset collection process started by dividing Arabic sentences into three groups. Each group contains around 500 unique sentences to be recorded by a group of signers (5 to 6 signers). Each sentence was targeted for at least five recordings by different signers. To reduce discrepancies across sentence groups, each sentence was first recorded by a certified ArSL expert and used as a reference sentence for other signers within the group.


To ensure diversity in recording conditions and signer appearance (e.g., different clothing, lighting, and backgrounds), the dataset was recorded by each signer in different sessions. Each session produced a single video containing 15–25 sentences. Capturing several sentences in one video eliminated the need to start and stop the recording for every sentence, thereby avoiding extra motion at sentence boundaries unrelated to the recorded sentence. We then segmented the videos into individual sentences using a semi-automated segmentation algorithm developed in this work. Specifically, expert signers were instructed to include a 2–3-second pause between consecutive sentences, allowing their hands to rest and move out of the camera frame. This facilitated automatic segmentation using the MediaPipe framework \cite{lugaresi2019mediapipe}, where hand-detection signals were used to identify extended temporal gaps corresponding to sentence boundaries. This strategy enabled accurate sentence-level separation for the majority of recordings, with only a few exceptions where signers did not fully adhere to the resting-hand guideline.

To analyze device heterogeneity, we tracked the types of mobile phones used for data collection. Fourteen signers recorded using iPhone devices (across multiple models), while two used Samsung and TECNO phones. The iPhone front-facing cameras had focal lengths ranging from 2.69 mm to 2.87 mm and apertures of f/2.2 or f/1.9. This variation in hardware contributes to the dataset’s realism, which reflects the range of devices typically used by members of the deaf and hard-of-hearing community. Moreover, the diversity of sign-capturing devices produced videos with resolutions ranging from 480p to 4K, providing a realistic and diverse benchmark for SLR. 

\subsection{Dataset Annotation and Verification}

The dataset was annotated by two sign language experts at two linguistic levels: the sign level through gloss sequences and the natural language level through Arabic sentences. Gloss annotations provide a sign-level representation of all signs within each sign language sentence while preserving the grammatical structure of the original sign language utterance. This type of annotation is essential for developing CSLR systems. In addition, we provide Arabic sentence annotations that capture the corresponding spoken-language meaning of each signed sentence, supporting the development of SLT systems. Furthermore, the availability of both gloss and Arabic annotations enables linguistic analysis and facilitates the study of the structural characteristics of ArSL.


The dataset underwent multiple verification stages to ensure recording, segmentation, and annotation quality. Since each signer recorded multiple sentences per session in a single video, the recordings were first segmented using the semi-automated segmentation procedure described earlier. The segmented clips were then reviewed to verify sentence boundaries and recording quality. Samples were re-recorded when the signer’s upper body was insufficiently visible or when the hands were not fully captured within the frame.

Finally, each segmented sign language sentence was checked against its intended reference sentence to ensure consistency in sign order and semantic content. In some cases, signers changed the order of signs, repeated certain gestures, or produced modified variants of the reference sentence. When such deviations were identified, the corresponding samples were flagged for re-recording or re-verification to maintain annotation quality and consistency across the dataset.


\subsection{Dataset Splits and Evaluation Protocols}
\label{section:benchmark_tasks}
 Following other sign language datasets~\cite{Zhou2021, alyami2025isharah}, we define two evaluation protocols: \textit{Signer-Independent} and \textit{Unseen-Sentences}.

\vspace{1mm}\noindent\textbf{Signer-Independent. }
For the signer-independent protocol, the dataset is partitioned such that the training, validation, and test sets comprise disjoint groups of signers. This setup enables the evaluation of a model’s ability to generalize to previously unseen signers. Under this split, {the training set contains 5,997 videos from 11 signers, the development set contains 1,500 videos from 3 signers, and the test set contains 1,500 videos from 4 signers.} While the sentence content overlaps across splits, the signers differ, ensuring that models must learn signer-invariant representations rather than memorizing individual signing styles.

\vspace{1mm}\noindent\textbf{Unseen-Sentences. }
For the unseen-sentences evaluation protocol, the objective is to evaluate whether a model can generalize to novel sentence compositions, rather than memorizing complete sentence patterns observed during training. In this setting, the sentences appearing in the development and test sets are not included in the training set. Although most individual glosses and Arabic words are already observed during training, they appear in novel combinations and linguistic contexts in the development and test splits. This setup evaluates a model’s ability to compose known lexical units into previously unseen sentences, which better reflects real-world SLR and SLT scenarios. Under this protocol, the training set contains 8,356 videos covering 1,377 unique gloss sentences and 1,383 unique Arabic sentences. The development and test sets contain 330 and 311 videos, corresponding to 49 and 50 unseen sentences, respectively.

\subsection{Dataset Statistics}

\begin{table}[]
    \centering
    \caption{Statistics of Isharah-Selfie dataset.}
    \label{tab:isharah_selfie_stats}
\resizebox{\linewidth}{!}{%
\footnotesize
\begin{tabular}{l|ccc|ccc|ccc|ccc}
\toprule
 & \multicolumn{6}{c|}{Sign Gloss (CSLR)} 
 & \multicolumn{6}{c}{Arabic Text (SLT)} \\
midrule
 & \multicolumn{3}{c}{Signer-independent} 
 & \multicolumn{3}{c|}{Unseen-sentences}
 & \multicolumn{3}{c}{Signer-independent} 
 & \multicolumn{3}{c}{Unseen-sentences} \\
 
 & Train & Dev & Test 
 & Train & Dev & Test
 & Train & Dev & Test 
 & Train & Dev & Test \\
\midrule

Videos 
 & 5997 & 1500 & 1500 & 8356 & 330 & 311
 & 5997 & 1500 & 1500 & 8356 & 330 & 311 \\

Unique Sentences 
 & 1476 & 1073 & 493 & 1377 & 49 & 50
 & 1482 & 1077 & 494 & 1383 & 49 & 50 \\

Avg.\ Sentence Length
 & 4.02 & 3.60 & 3.28 & 3.72 & 5.58 & 4.96
 & 3.50 & 3.21 & 2.94 & 3.36 & 3.35 & 3.39 \\

Vocab Size 
 & 961 & 814 & 497 & 959 & 113 & 129
 & 2133 & 1689 & 889 & 2037 & 133 & 135 \\

Singletons (\%) 
 & 40.37 & -- & -- & 42.65 & -- & --
 & 64.14 & -- & -- & 64.65 & -- & -- \\

OOV (\%) 
 & -- & 0.0 & 0.0 & -- & 1.77 & 0.78
 & -- & 0.0 & 0.0 & -- & 41.35 & 30.37 \\

\bottomrule
\end{tabular}
}
\end{table}

The \textit{Isharah-Selfie} dataset includes 8,997 sign language videos spanning 12.5 hours, performed by 16 signers across 1,493 unique sentences. The participants include 8 Deaf, 6 hearing, and 2 hard-of-hearing signers; 11 male and 5 female signers; and age groups ranging from 21 to 50 years. The recordings were self-captured using front-facing smartphone cameras, resulting in diverse capture conditions, including different resolutions, camera quality, close-range framing, partial body visibility, lighting changes, background clutter, and camera motion. The most common resolutions are 1920$\times$1080 (38\%) and 3840$\times$2160 (25\%), with additional recordings ranging from 480p to 4K.

To characterize the one-handed nature of the dataset, we analyzed frame-level hand presence using MediaPipe detections. As shown in Figure~\ref{fig:frame_state_distribution}, single-hand frames dominate the dataset: right-hand-only frames account for 49.1\%, left-hand-only frames for 16.6\%, and both-hand frames for only 2.7\%. Frames with no detected hands account for 31.6\%, reflecting natural pauses, transitional movements, hands moving outside the camera field of view, and occasional detection failures. This distribution confirms that \textit{Isharah-Selfie} is structurally different from conventional two-handed CSLR benchmarks.

The dataset also exhibits non-negligible camera motion due to handheld recording. Using a global affine model estimated between consecutive frames, we found that most videos exhibit low-to-moderate motion, indicating that the camera is typically non-stationary rather than fixed. This adds realistic visual variability compared with controlled CSLR datasets recorded using static cameras.

Table~\ref{tab:isharah_selfie_stats} summarizes the CSLR and SLT statistics under the signer-independent and unseen-sentences protocols. Overall, the annotations consist of short sentence-level sequences, with average gloss lengths ranging from 3.28 to 5.58 tokens and average Arabic text lengths ranging from 2.94 to 3.50 tokens. In the signer-independent setting, the training, development, and test splits contain 5,997, 1,500, and 1,500 videos, respectively. In the unseen-sentences setting, the training set contains 8,356 videos covering 1,377 unique gloss sentences and 1,383 unique Arabic sentences, while the development and test sets contain 330 and 311 videos from 49 and 50 unseen sentences, respectively. Gloss-level out-of-vocabulary (OOV) rates remain low in the unseen-sentences split (1.77\% dev and 0.78\% test), whereas Arabic text has higher OOV rates (41.35\% dev and 30.37\% test), highlighting the greater lexical variability of the SLT task.


\section{Benchmark Experiments}
\label{sec:experiments}

\vspace{1mm}\noindent\textbf{Baseline Models. }
We evaluate \textit{Isharah-Selfie} dataset using several state-of-the-art (SOTA) models for both CSLR and SLT tasks. For CSLR, we adopt TLP \cite{hu2022temporal}, Swin-MSTP \cite{ALYAMIswin}, USTM \cite{hasanaath2025ustm}, SlowFastSign \cite{slowfast2024}, and CorrNet+ \cite{hu2024corrnet+} models. The TLP model was proposed for CSLR, and it improves temporal efficiency through compact feature representations. Swin-MSTP integrates Swin Transformers with multi-scale CNNs for robust spatio-temporal modeling, while SlowFastSign captures multi-rate motion dynamics via dual temporal pathways. CorrNet+ explicitly models inter-frame relationships by constructing human body trajectories with correlation and identification modules, followed by temporal attention. We further include pose-based CSLR baselines, namely MSLR-Pose \cite{luqman2025signeval} and VIPL-SLP \cite{Min_2025_ICCV} models. MSLR-Pose leverages 2D hand keypoints processed through Transformer encoder layers with positional encoding and temporal pooling for sequential modeling. VIPL-SLP employs a hybrid CNN and GCN architecture that integrates group-wise skeletal modeling and temporal convolutions. For SLT, we consider two recent frameworks GFSLT-VLP \cite{Zhou_2023_ICCV} and SpaMo \cite{hwang2025efficient}. GFSLT-VLP performs direct video-to-text translation using contrastive and masked-text learning within a vision-language pretraining framework, whereas SpaMo leverages spatial and motion-aware visual encoders combined with large language models to directly generate translations from sign videos, enhanced by a visual-text alignment warm-up to improve cross-modal consistency.

\vspace{1mm}\noindent\textbf{Implementation Details. }
For all benchmark models, we adhered to the original training and inference configurations as described in their respective works, with a few necessary modifications. In the SLT experiments, we replaced the LLMs used in GFSLT-VLP and SpaMo with mT0-xl, as the originally used LLMs did not support Arabic. Additionally, for pose-based CSLR methods, specifically VIPL-SLP, we substituted the GCN with an MLP, which empirically yielded better performance.
All experiments were conducted on a single NVIDIA RTX A6000 GPU with 48GB of VRAM. 



\vspace{1mm}\noindent\textbf{Evaluation Metrics. }
We evaluate CSLR using word error rate (WER), which measures the edit distance between predicted and ground-truth gloss sequences; lower WER indicates better recognition. For SLT, we report BLEU-1 to BLEU-4 to measure n-gram overlap between generated and reference Arabic sentences, along with ROUGE-L to assess longest common subsequence similarity.


\begin{table}[t]
       \centering
    \caption{Comparison of RGB-based and pose-based CSLR models' performance in WER (\%).}
    \label{tab:slr_rgb_pose_combined}
\resizebox{.6\linewidth}{!}{%
\footnotesize{
    \begin{tabular}{l l cc cc}
      \toprule
       &
      \multirow{2}{*}{Model} &
      \multicolumn{2}{c}{Signer-Independent} &
      \multicolumn{2}{c}{Unseen-Sentences} \\
      \cmidrule(lr){3-4}\cmidrule(lr){5-6}
      & & Dev & Test & Dev & Test \\
      \midrule

      \multirow{5}{*}{RGB}
      & TLP~\cite{hu2022temporal}
      & 55.87 & 55.30 & 74.51 & 75.92 \\

      & Swin-MSTP~\cite{ALYAMIswin}
      & 84.61 & 85.01 & 85.49 & 88.07 \\

      & USTM~\cite{hasanaath2025ustm}
      & 85.00 & 87.33 & 86.31 & 87.72 \\

      & CorrNet+~\cite{hu2024corrnet+}
      & \textbf{40.1} & \textbf{39.83} & \textbf{68.64} & \textbf{68.29} \\

      & SlowFastSign~\cite{slowfast2024}
      & 70.36 & 66.63 & 83.91 & 84.05 \\

      \midrule

      \multirow{2}{*}{Pose}
      & MSLR-Pose~\cite{luqman2025signeval}
      & 58.48 & 60.98 & 83.91 & 84.05 \\

      & VIPL\_SLP~\cite{Min_2025_ICCV}
      & 69.10 & 67.09 & 80.69 & 79.44 \\

      \bottomrule
    \end{tabular}
    }}
\end{table}

\subsection{Results and Discussion}
Table~\ref{tab:slr_rgb_pose_combined} reports the CSLR results under the signer-independent and unseen-sentence protocols. In the signer-independent setting, CorrNet+ achieves the lowest WER on both the development and test sets, indicating that explicit modeling of inter-frame correlations is beneficial under selfie-style recording conditions. In contrast, models such as Swin-MSTP, USTM, and SlowFastSign obtain substantially higher WER, suggesting that strong spatio-temporal backbones alone are not sufficient when videos contain close-range viewpoints, partial body visibility, camera motion, and one-handed articulation.

The unseen-sentences setting is more challenging for all CSLR models. CorrNet+ remains the strongest RGB-based method, but its WER increases considerably compared with the signer-independent setting. This confirms that sign language sentence recognition requires not only signer-invariant representations, but also the ability to generalize to novel sentence compositions. The large gap between signer-independent and unseen-sentences performance suggests that current CSLR models are not only sensitive to visual variability but also rely heavily on sentence-level patterns observed during training, limiting their ability to compose known glosses into novel utterances.

Pose-based CSLR methods also show limited performance compared with the best RGB-based model. Although pose representations provide a compact abstraction of body motion, selfie-style signing introduces several sources of noise, including hands frequently moving outside the frame, unreliable keypoints from the non-signing arm, and reduced pose stability caused by close-range camera motion. These issues are consistent with the hand-presence analysis in Figure~\ref{fig:frame_state_distribution}, where single-hand frames dominate the dataset and both-hand detections are rare. As a result, pose-based models must distinguish meaningful signing motion from missing or noisy keypoints, which remains difficult in this setting.

Table~\ref{tab:slt_res} presents the SLT results. Under the signer-independent protocol, both GFSLT-VLP and SpaMo achieve reasonable BLEU-1 and ROUGE-L scores, with SpaMo performing better on the test set. However, both models degrade sharply under the unseen-sentences protocol, especially for BLEU-3 and BLEU-4. This indicates that the models can often recover individual words or short fragments but struggle to generate longer, correct word sequences for unseen sentence compositions.

The n-gram coverage analysis in Table~\ref{tab:ngram_coverage} helps explain this degradation. While unigram coverage in the unseen-sentences splits remains moderate, bigram coverage drops substantially, and trigram and four-gram coverage are almost absent. Since BLEU-3 and BLEU-4 depend on exact higher-order n-gram matches, the low coverage of longer word sequences makes high scores difficult to achieve. These results show that \textit{Isharah-Selfie} is a challenging benchmark for compositional generalization in both CSLR and SLT. These findings suggest that selfie-style CSLR requires models that are robust to missing or unreliable hand cues, asymmetric one-handed motion, and compositional generalization beyond sentence patterns seen during training.




\begin{table*}[h]
\tiny
  \centering
  \caption{Performance comparison of SLT models. }
  \label{tab:slt_res}
  \resizebox{\textwidth}{!}{%
    \begin{tabular}{ll ccccc ccccc}
      \toprule
\multicolumn{2}{c}{} &
\multicolumn{5}{c}{\textbf{Dev}} &
\multicolumn{5}{c}{\textbf{Test}} \\
\cmidrule(lr){3-7}\cmidrule(lr){8-12}
\textbf{Model} & \textbf{Split } &
\textbf{BLEU1} & \textbf{BLEU2} & \textbf{BLEU3} & \textbf{BLEU4} & \textbf{ROUGE-L} &
\textbf{BLEU1} & \textbf{BLEU2} & \textbf{BLEU3} & \textbf{BLEU4} & \textbf{ROUGE-L} \\
\midrule

\multirow{3}{*}{\textbf{GFSLT-VLP}~\cite{Zhou_2023_ICCV}} 
& Signer-Independent  & 43.68 & 32.22 & 28.46 & 26.30 & 55.11 & 30.74 & 13.08 & 9.34 & 7.67 & 45.11 \\
& Unseen-Sentences     & 23.44 & 1.96 & 0.60 & 0.31 & 38.99 & 23.01 & 5.35 & 1.17 & 0.51 & 20.7 \\
\midrule

\multirow{3}{*}{\textbf{SpaMo}~\cite{hwang2025efficient}}
& Signer-Independent  & 42.92 & 31.26 & 27.27 & 24.89 & 55.88 & 33.52 & 17.43 & 13.47 & 11.40 & 47.72\\
& Unseen-Sentences     & 22.35 & 1.32 & 0.45 & 0.24 & 37.97 & 21.15 & 2.93 & 0.77 & 0.37 & 37.77 \\

\bottomrule
    \end{tabular}%
  }
\end{table*}

\begin{table*}[h]
\centering
\footnotesize
\caption{N-gram coverage of words from the unseen-sentences dev and test splits in the training set, broken down by the dominant hand assignment of each video. For each n-gram, we report the number of n-grams that appear in training under the same hand label (Covered), those that do not (Uncovered), and the overall coverage rate (Coverage \%).}
\label{tab:ngram_coverage}
\renewcommand{\arraystretch}{1.2}
\setlength{\tabcolsep}{4pt}

\begin{tabular}{l ccc ccc}
\toprule
& \multicolumn{3}{c}{Dev} & \multicolumn{3}{c}{Test} \\
\cmidrule(lr){2-4}\cmidrule(lr){5-7}
\textbf{N-gram} 
& Covered & Uncovered & Coverage (\%)
& Covered & Uncovered & Coverage (\%) \\
\midrule
1-gram & 711 & 396 & 64.2\% & 804 & 251 & 76.2\% \\
2-gram & 179 & 598 & 23.0\% & 217 & 527 & 29.2\% \\
3-gram & 20  & 427 & 4.5\%  & 18  & 415 & 4.2\% \\
4-gram & 0   & 180 & 0.0\%  & 0   & 203 & 0.0\% \\
\bottomrule
\end{tabular}
\end{table*}

\section*{Broader Impact}

\textit{Isharah-Selfie} supports the development of more realistic SLR and SLT systems for smartphone-based communication. Although one-handed selfie signing is common in mobile communication among Deaf individuals, it remains largely understudied in existing CSLR and SLT benchmarks for both ArSL and other sign languages. By capturing this setting, \textit{Isharah-Selfie} provides a benchmark for studying CSLR and SLT under everyday mobile conditions, including one-handed articulation, close-range viewpoints, partial body visibility, camera motion, and unconstrained environments.

The dataset may improve accessibility for the Deaf and hard-of-hearing community in Saudi Arabia and the wider Arab region by encouraging technologies that reflect natural communication practices. The dataset will be released exclusively for non-commercial research use under a signed usage agreement, as it contains identifiable videos and covers sensitive domains such as healthcare, legal services, and emergency communication. Models trained on this dataset should not be used for surveillance, biometric identification, or high-stakes deployment without additional validation.

\section{Limitations}

\textit{Isharah-Selfie} has several limitations. First, the signer demographics are imbalanced, with male signers and the 31--40 age group overrepresented, which may affect generalization to underrepresented groups. Second, the dataset focuses on ArSL, so findings may not directly transfer to other sign languages or signing communities. Third, although annotations were prepared from reference recordings and verified across signers, minor inconsistencies may remain when signers produced valid variations of the same sentence. Fourth, selfie-based one-handed signing can compress or modify inherently bimanual signs, making some signs visually ambiguous. {\color{black}Fifth, the dataset mainly contains short utterances and is therefore less suitable for evaluating long-term dependencies in extended discourse.} Finally, the reported baselines were evaluated using single runs, and pose-based methods may be affected by keypoint noise under close-range mobile recording conditions.


\section{Conclusion}
\label{sec:conclusion}
We proposed \textit{Isharah-Selfie}, a large-scale dataset for CSLR and SLT under realistic selfie-style recording conditions. The dataset offers 8,997 front-camera smartphone videos from 16 signers and 1,493 unique sentences, with both gloss-level and Arabic sentence annotations. Unlike controlled benchmarks, \textit{Isharah-Selfie} captures everyday mobile signing with one-handed articulation, partial body visibility, close-range framing, camera motion, diverse backgrounds, variable lighting, and heterogeneous device resolutions.
We also benchmarked \textit{Isharah-Selfie} dataset using SOTA RGB-based and pose-based CSLR and SLT models. The obtained results reveal that \textit{Isharah-Selfie} is a challenging benchmark for both CSLR and SLT under selfie-style signing conditions. 
All CSLR models experience substantial performance degradation under the unseen-sentences protocol, indicating limited compositional generalization to novel sentence structures. Pose-based methods perform worse than the best RGB-based approaches due to noisy or missing keypoints caused by one-handed signing, partial hand visibility, and camera motion. Similarly, SLT models achieve reasonable performance under signer-independent evaluation but degrade sharply on unseen sentences, particularly for higher-order BLEU metrics.
These findings highlight the need for CSLR and SLT models that are robust to selfie-style recording challenges and capable of stronger compositional generalization beyond sentence patterns observed during training. 
By releasing \textit{Isharah-Selfie}, we aim to support the development of practical, inclusive, and smartphone-ready sign language technologies for real-world communication.

\section*{Acknowledgements and Disclosures}

\paragraph{Funding.}
The authors would like to acknowledge the support received from the Saudi Data and AI Authority (SDAIA) and King Fahd University of Petroleum and Minerals (KFUPM) through the SDAIA-KFUPM Joint Research Center for Artificial Intelligence.

\paragraph{Competing interests.}
The authors declare that they have no competing interests. No third-party support was received during the last 36 months for the work submitted, and the authors have no financial relationships outside the submitted work that could be perceived as influencing it.


%

\bibliographystyle{unsrt}  
\bibliography{main}

\end{document}